%% file: root.tex
\documentclass[letterpaper, 10 pt, conference]{ieeeconf}  

\IEEEoverridecommandlockouts                              

\usepackage{graphics} 
\usepackage{amsmath,amsfonts}
\usepackage{algorithm}
\usepackage[noend]{algpseudocode}
\usepackage{array}
\usepackage{textcomp}
\usepackage{stfloats}
\usepackage{url}
\usepackage{verbatim}
\usepackage{cite}
\usepackage{xcolor}
\usepackage{todonotes}
\usepackage{float} 
\usepackage{graphicx}
\usepackage{subcaption}
\usepackage{multirow} 
\usepackage{array}

\DeclareUnicodeCharacter{221E}{}

\newcommand{\abbv}{\text{HJARL}}

\title{\LARGE \bf
Learning Robust Policies via Interpretable Hamilton-Jacobi Reachability-Guided Disturbances}

\author{Hanyang Hu$^{1}$, Xilun Zhang$^{2}$, Xubo Lyu$^{1}$, Mo Chen$^{1}$
\thanks{*This work was supported by the Canada CIFAR AI Chairs and NSERC Discovery Grants Programs.}
\thanks{$^{1}$Hanyang Hu, Xubo Lyu, and Mo Chen are with Simon Fraser University, Burnaby, BC V5A 1S6, Canada. $^{2}$Xilun Zhang is with Carnegie Mellon University, Pittsburgh, PA 15213, USA.}%
\thanks{We thank Satvik Garg, Wenxiang He, and Hanjie Liu for their help with real-world experiments.}%
}

\begin{document}

\maketitle
\thispagestyle{empty}
\pagestyle{empty}

\begin{abstract}
Deep Reinforcement Learning (RL) has shown remarkable success in robotics with complex and heterogeneous dynamics.
However, its vulnerability to unknown disturbances and adversarial attacks remains a significant challenge. 
In this paper, we propose a robust policy training framework that integrates model-based control principles with adversarial RL training to improve robustness without the need for external black-box adversaries. 
Our approach introduces a novel Hamilton-Jacobi reachability-guided disturbance for adversarial RL training, where we use interpretable worst-case or near-worst-case disturbances as adversaries against the robust policy. 
We evaluated its effectiveness across three distinct tasks: a reach-avoid game in both simulation and real-world settings, and a highly dynamic quadrotor stabilization task in simulation. 
We validate that our learned critic network is consistent with the ground-truth HJ value function, while the policy network shows comparable performance with other learning-based methods.
\end{abstract}

\input{sections/Introduction}
\input{sections/Related_Work}
\input{sections/Preliminaries}
\input{sections/Method}
\input{sections/Numerical_Simulation}
\input{sections/Experiments}
\input{sections/Conclusion}

\addtolength{\textheight}{0cm}   

\bibliographystyle{ieeetr}
\bibliography{ref.bib}

\end{document}

%% file: sections/Introduction.tex
\section{Introduction}

Deep Reinforcement Learning has emerged as a powerful tool in robotics, particularly within highly dynamic environments\cite{lyu2020ttr, zhang2024wococo, cheng2024extreme, wang2023guardians}.
However, its trained policies may fail when the simulation to real-world gap is large due to modeling errors and unknown disturbances \cite{xu2022trustworthy,huang2023went}. Consequently, developing a Robust RL (RRL) policy is crucial to prevent catastrophic policy failures during deployment\cite{8967695}.

To address the issues of model mismatches and unforeseen disturbances, recent advances in RRL focus on maximizing worst-case performance across various uncertainties. 
One widely recognized approach within RRL is Robust Adversarial Reinforcement Learning (RARL) \cite{pinto2017robust} which mitigates environmental mismatches by treating them as adversarial perturbations. 
RARL conceptualizes the problem as a two-player zero-sum game, in which the protagonist aims to develop a robust policy across different environments, while the adversary introduces the perturbation policy to challenge the protagonist. However, such learning-based adversarial agents often lack theoretical interpretability and may generate implausible disturbances \cite{brunke2022safe}. 
In this paper, we aim to develop a robust adversarial RL framework that enables more interpretable and verifiable solutions for adversary and protagonist policies by leveraging robust control theory, which also achieves comparable performances with state-of-the-art under various disturbances.

\begin{figure}[t]
\centering
\includegraphics[width=0.48\textwidth]{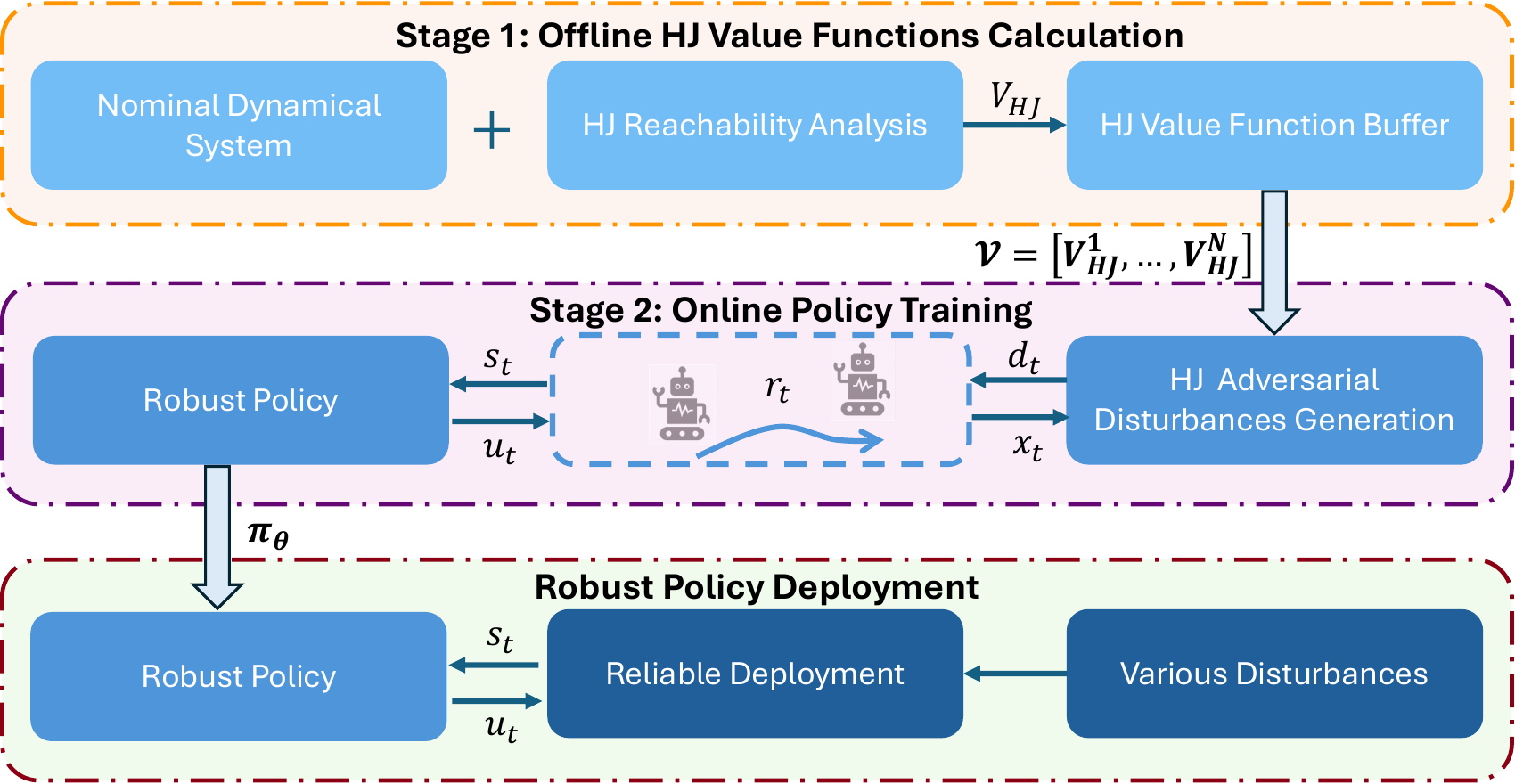}
\caption{\abbv\ computes the HJ value functions offline and uses them to generate adversarial disturbances during the online training. The trained robust policy is then deployed to handle various disturbances.}
\label{pipeline}
\end{figure}

Hamilton-Jacobi (HJ) reachability analysis is a robust optimal control method grounded in game theory. 
It treats disturbances as adversaries and provides robust minimax optimal policies for both agents, regardless of linear or nonlinear dynamics \cite{chen2018hamilton}. 
HJ reachability solves a differential game using the Hamilton-Jacobi-Isaacs (HJI) equation. 
The HJ value function has proven to be the viscosity solution to this HJI equation and can be used to calculate optimal controls and disturbances for opposing parties \cite{fisac2015reach}. 
As a model-based approach, HJ reachability offers physically interpretable protagonist actions and adversarial disturbances. 
Its robustness is demonstrated by the guaranteed outcomes when initial states lie within a specific region of the value function. 
However, due to computational limitations, accurate solutions to the HJ reachability become intractable if the state space is higher than six dimensions (6D) \cite{bui2022optimizeddp}. 
To mitigate this curse of dimensionality issue, we utilize the nominal dynamics for the HJ computation to circumvent the full high-dimensional dynamics and focus on low-dimensional tractable solutions instead of using learning-based approximation tools.

In this paper, we propose a novel method \abbv, short for \textbf{H}amilton-\textbf{J}acobi-guided \textbf{A}dversarial \textbf{RL} training as shown in Fig.\ref{pipeline}. 
Our method differs from other learning-based robust adversarial RL methods, opting to generate adversarial disturbances through HJ reachability directly instead of learning adversarial policy networks. 
By leveraging nominal dynamical systems, we compute solvable HJ value functions offline in stage 1. This enables the generation of worst-case or near-worst-case disturbances during online training (stage 2), thus improving policy robustness.
To smooth the learning process, we utilize the Boltzmann distribution to gradually increase the strength of the adversary to avoid overly strong adversaries in the initial training phase.
The model-based nature of HJ reachability guarantees that the generated adversarial actions stay within a physically feasible bound.
\abbv\ obtains a critic network that aligns with the HJ value function, providing an approximation to the HJ  guaranteed reachable region, and achieves comparable task performances with other learning-based methods.
It also requires less parameter tuning and achieves faster convergence than training adversarial policy networks.
Our contributions are as follows:
\begin{enumerate}
    \item Propose a novel method to obtain robust control policies that integrates the adversarial RL framework with HJ reachability using nominal dynamical systems.
    \item Demonstrate the robustness and the consistency to the HJ reachability analysis on the low-dimensional dynamical system and achieve comparable task performances against other baseline methods on the high-dimensional dynamical system. 
    \item Validate our robust control policy in a real-world one vs. one reach-avoid game using TurtleBot3 robots.
\end{enumerate}

%% file: sections/Related_Work.tex
\section{Related Work}

\textbf{Robust Adversarial RL.} Motivated by $H_\infty$ control, pioneering works by \cite{morimoto2005robust, pinto2017robust, oikarinen2021robust} introduced the concept of RARL which trains robust control policies against learned adversarial disturbances within the game theory framework. 
They formulate zero-sum Markov games as a Markov Decision Process (MDP)\cite{perolat2015approximate, puterman1990markov}, and generally learn a universal function approximator to approximate adversarial policies. 
A range of works have been proposed to address the limitations of the RARL framework. 
Xinlei et al. proposed a risk-averse RARL algorithm with a risk-averse protagonist and risk-seeking adversary to account for the probability of catastrophic events \cite{pan2019risk}. 
Kaiqing et al. demonstrated that RARL cannot guarantee training stability or convergence, even in linear quadratic cases \cite{zhang2020stability}.  
Reddi et al. proposed an entropy-regularization-based RARL method to simplify the saddle point optimization of the original algorithm \cite{reddi2023robust}. 
To address the problem that employing a single adversarial network could lead to biased adversary generation, \cite{vinitsky2020robust} advocated the use of multiple adversarial networks along with a performance ranking system to ensure more effective training of robust policies. 
Rather than formulating adversarial training as a zero-sum game, \cite{huang2022robust} models RRL training as a Stackelberg game, where the adversarial agent is adaptively regularized to improve the stability of the training.
Although these methods have shown promising results under the RL framework, they lack a physically interpretable analysis of the adversary, and the trade-off between training convergence and robustness still remains challenging. 

\textbf{Hamilton-Jacobi Reachability.} 
$H_\infty$-control theory was first proposed for the design of the worst-case disturbance controller considering the linear dynamical system in the frequency domain \cite{zames1981feedback,francis1987lecture, aliyu2011nonlinear}. 
Then, \cite{doyle1988state, van1991state, van19922} extended the $H_\infty$-control theory to nonlinear dynamical systems in the state-space domain, and this resulting minimax optimization problem naturally connects to game theory and differential game \cite{bacsar1998dynamic, bacsar2008h}.  
HJ reachability analysis is a model-based robust optimal control method under the framework of differential games \cite{chen2018hamilton}. 
\cite{gong2024robust, yang2024robust, wang2024providing} provide robust control frameworks grounded in the HJ value function and the corresponding Backward Reachable Tubes (BRT). 
Due to the curse of dimensionality, 
numerical solutions to HJ reachability are limited to 6D problems using the latest toolbox \cite{bui2022optimizeddp}.  
To circumvent this constraint, \cite{chen2016exact} introduced a dimensionality reduction technique. 
In addition to these numerical computation methods, researchers have used neural networks to approximate the HJ value function.  
Kai-Chieh et al. utilized a time-discounted Safety Bellman Equation with adversarial disturbances and built a new scheme to provide certiﬁed safe actions \cite{fisac2019bridging, hsu2023isaacs}. 
Somil et al. proposed the DeepReach method that uses sinusoidal networks as a partial differential equation solver to approximate the HJ value function after long offline training \cite{bansal2021deepreach}. 
Although these learning-based approximations can be used for higher-dimensional problems, their correctness and accuracy are hard to verify.

Unlike previous work on adversarial training, \abbv\ provides a physically theoretical bound for the disturbance, generating the worst or near-worst adversarial disturbances incrementally during training.
This approach also provides an interpretable viewpoint on the robustness of the learned critic network compared to previous methods while achieving comparable task performances. 

%% file: sections/Preliminaries.tex
\section{Preliminaries}

\subsection{Hamilton-Jacobi Reachability Analysis}
HJ reachability analysis constitutes a model-based approach to robust optimal control. 
The following ordinary differential equations govern the \textbf{nominal dynamical system}:
\begin{equation}
\label{whole_dynamics}
    \dot{x} = f(x, u, d), \quad x(0) = x_0
\end{equation}
where $x\in \mathbb{R}^n$ is the state we use for the HJ value function computation, 
$f$ represents the nominal dynamical system, which may either encompass the complete dynamics or only the partial dynamics relevant to the states of interest,
particularly in cases where computational intractability is a concern, 
$u \in \mathcal{U}$ is the control input, 
$d \in \mathcal{D}$ is the disturbance to the system,
and $x_0$ is the initial state. 
The sets $\mathcal{U}$ and $\mathcal{D}$ are the sets of measurable functions.

HJ reachability analysis resides at the intersection of optimal control and differential games, enabling adversarial parties in a minimax game to achieve their optimal control objectives respectively \cite{chen2018hamilton}. 
In our setting, we refer to the disturbance of the control system as the adversary in a general differential game. 
The goal of the control input is to push the system to the target set $R$. 
The target set $R$ can then be defined by an implicit surface function $l(x): \mathbb{R}^n \rightarrow \mathbb{R}$ such that $R =\left\{x \in \mathbb{R}^n \mid l(x) \leq 0\right\}$ using the level set method. 
Similarly, the avoid set $A$ that stands for the constraints can also be formulated by the function $g(x): \mathbb{R}^n \rightarrow \mathbb{R}$ such that $A =\left\{x \in \mathbb{R}^n \mid g(x) > 0\right\}$. 
Given these expressions, let the value function  $V_{HJ}: \mathbb{R}^n \times[-T, 0] \rightarrow \mathbb{R}$ be the viscosity solution to the HJI partial differential equation within time $[-T, 0]$ \cite{crandall1983viscosity}:
\begin{equation}
\label{HJI}
\begin{aligned}
\max \{  & \min \left\{\frac{\partial V_{HJ}}{\partial t} + H \left(x, p\right), l(x)-V_{HJ}(x, t) \right\}, \\
& \, \left.g(x)-V_{HJ}(x, t)\right\}=0, \quad t \in[-T, 0]
\end{aligned}
\end{equation}
where the optimal Hamiltonian $H$ is calculated as:
\begin{equation*}
H(x, p)=\min _{u \in \mathcal{U}} \max _{d \in \mathcal{D}} p^T f(x, u, d)
\end{equation*}
where $p = \frac{\partial V_{HJ}}{\partial x}$. 
If the avoid set $A$ does not exist, the $g(x)$ can be set to negative infinite.
With the solved value function $V_{HJ}$, one can construct the BRT where the value of the state is negative. Starting from the BRT, the agent will reach the target set under optimal control regardless of the disturbance within the time interval.

Since there are no general analytical solutions for Eq.\eqref{HJI} mostly, one must rely on numerical solutions.
These methods hinge on discretizing the state space and employing dynamic programming iterations. 
Consequently, computational and spatial complexity grows exponentially with the expansion of state quantities. 
Numerical computation tools like the OptimizedDP \cite{bui2022optimizeddp} leverage contemporary computational capabilities to effectively tackle Eq. \eqref{HJI} with grid resolutions of reasonable quality, extending to six dimensions.

If we take the time interval $T\rightarrow \infty$, we will obtain control policies that are independent of the time $t$. 
Then, the optimal control and the worst disturbance can be calculated as \cite{huang2011differential}:
\begin{equation}
\label{eq: optimal_strategies}
\begin{aligned}
& u^*(x)=\arg \min _{u \in \mathcal{U}} \max _{d \in \mathcal{D}} p(x)^\top f(x, u, d) \\
& d^*(x)=\arg \max _{d \in \mathcal{D}} p(x)^\top f\left(x, u^*, d\right)
\end{aligned}
\end{equation}

In this way, as for the nominal dynamics Eq.\eqref{whole_dynamics}, given the pre-computed value function and the agent's current state, the optimal adversarial control working as the worst disturbance to the protagonist is obtained by Eq. \eqref{eq: optimal_strategies}.

\subsection{Robust Adversarial RL}
The adversarial RL aims to learn a policy that maximizes the expected rewards under the worst disturbances. 
The objective function under the disturbance is as follows :
\begin{equation}
\label{eq: adversarial RL}
\begin{gathered}
\max _\theta \min _{\phi} \mathbb{E}\left[\sum_{t=0}^T \gamma^t r\left(s_t, u_t, d_t \right) \mid \pi_\theta, \pi_{\phi}\right]
\end{gathered}
\end{equation}
where $\pi_{\theta}$ is the goal robust policy parameterized by $\theta$,
$u_t \sim \pi_{\theta}(s_t)$ is the action sampled from the policy given the discrete state $s_{t}$ at the time step $t$, 
$\pi_{\phi}$ is the learnable adversary policy parameterized by $\phi$, 
and $d_t \sim \pi_{\phi}(s_t)$ is the disturbance sampled from this adversarial policy.

%% file: sections/Method.tex
\section{Methodology}

\subsection{Problem Formulation}
\textbf{Markov Decision Process.}  
We consider a MDP defined by the 6-tuple $(\mathcal{S}, \mathcal{A}, P, r, \gamma, \mu_0)$, where $\mathcal{S}$ is the set of states, and $\mathcal{A}$ is the set of actions, which can be continuous or discrete. 
The function $P: \mathcal{S} \times \mathcal{A} \rightarrow \Delta(\mathcal{S})$ represents the transition probability, where $\Delta(\mathcal{S})$ denotes the distribution over the state space $\mathcal{S}$. 
The reward function is defined as $r: \mathcal{S} \times \mathcal{A} \rightarrow \mathbb{R}$, $\gamma \in [0, 1)$ is the discount factor and $\mu_0: \mathcal{S} \rightarrow \mathbb{R}_{+}$ represents the initial state distribution. 
The objective in RL is to find a policy $\pi_\theta: \mathcal{S} \times \mathcal{A} \rightarrow \mathbb{R}$, parameterized by $\theta$, that maximizes the expected return $\mathbb{E}_{\tau \sim \pi_\theta} \left[R(\tau)\right]$, where the return $R(\tau)$ is defined as:$R(\tau) := \sum_{t=0}^{T-1} \gamma^t r(s_t, a_t)$, and $\tau$ denotes the trajectories sampled using the policy $\pi_\theta$.

\subsection{HJ Reachability-Guided Adversarial Training}
\abbv\ adds adversarial disturbances generated by the HJ value function $V_{HJ}$ directly to the agent's actions. 
To ensure the generalization of the learned robust policy, $V_{HJ}$ is calculated through a sequence of increasing upper bounds of disturbance. 
We denote nominal dynamical systems as $[f^1, \ldots, f^i, \ldots, f^N]$ where the upper bound of the disturbances increases uniformly. 
These nominal dynamical systems then construct a list of Hamiltonians $[H^1(x, p^1), \ldots, H^i(x, p^i), \ldots,  H^N(x, p^N)]$, and finally obtain a HJ value function buffer: $\mathcal{V} = [ V^1_{HJ}, \ldots, V^i_{HJ}, \ldots, V^N_{HJ}]$.
The upper bound of the disturbance is selected based on physically feasible significance and keeps the same order of magnitude as the allowed control input. 
Given the certain value function $V^i_{HJ}$, the current action of the protagonist $u_t$, and the current state of the agent $s_t$, we can compute the disturbance $d_t$ from the corresponding deterministic disturbance-generated policy $d^i$ based on the $f^i$:
\begin{equation}
    d_t = d^i(s_t|V^i_{HJ}) = \left.\arg \max \left(\frac{\partial V_{H J}^i}{\partial x}\right)^{\top}\right|_{x=x_t} f^i\left(x_t, u_t, d_t\right)
\label{disturbance}
\end{equation}
where $x_t$ is the nominal state which is the same or a part of the full state $s_t$ at the time step $t$.

\abbv\ aims to learn a policy that maximizes the expected rewards in the presence of disturbances generated by Eq.\eqref{disturbance}. 
The resulting optimization problem under these disturbances is formulated as follows:
\begin{equation}
\label{eq: our_goal}
\begin{gathered}
\max _{\pi_\theta} \mathbb{E}_{i \sim B(1, N)}\left[\sum_{t=0}^T \gamma^t r\left(s_t, u_t, d_t \right) \mid \pi_\theta, d^{i}\right] 
\end{gathered}
\end{equation}
where $\pi_{\theta}$ is the task policy parameterized by $\theta$, 
$u_t \sim \pi_{\theta}(s_t)$ is the action sampled from the policy. 
The Boltzmann distribution $ i \sim B(1, N)$ is used to obtain a smooth curriculum learning process by sampling a value function $V^i_{HJ}$ from the HJ value function buffer $\mathcal{V}$ for each episode.

Algorithm \ref{alg: HARLT} outlines the \abbv\ in detail. 
In the first stage (line 1 to line 5), the algorithm precomputes value functions that account for varying levels of disturbances in the environment. 
Specifically, we define the disturbance levels and calculate the corresponding value function $V_i$ for the nominal dynamical system $f$. 
The calculation is performed with the OptimizedDP numerical toolbox \cite{bui2022optimizeddp}.
Upon completion of these computations, the resulting value functions are aggregated into a value function buffer $\mathcal{V}$. 
In the second stage (line 6 to the end), we perform online adversarial RL training, intending to train a policy $ \pi_\theta$ that can effectively handle disturbances characterized by the value functions generated in the first stage. 
During each rollout, we sample a value function $V^i_{HJ}$ from the value function buffer $\mathcal{V}$ at the beginning of each trajectory $\tau_{j}$; the corresponding disturbances $d^{i}$ are generated based on the sampled value function $V^i_{HJ}$ and are applied throughout the trajectory. 
After collecting \textit{M} trajectories, we choose Proximal Policy Optimization (PPO) \cite{schulman2017proximal} to update our policy.
This iterative update process continues until the policy converges.

\begin{algorithm}[tb]
   \caption{HJ Reachability Guided Adversarial Training}
   \label{alg: HARLT}
\begin{algorithmic}[1]
   
    \State \textbf{Value Function Generation with HJ Reachability}
    \State Initialize the value function buffer $\mathcal{V}$, the number of the disturbance levels $N$, the nominal dynamical system $f$;
   \For{$i = 1, \ldots, \textit{N}$}
        \State solve for $V^i_{HJ}$ based on $f^i$ according to Eq.\eqref{HJI};
        \State collect the value function $\mathcal{V} \leftarrow \mathcal{V} \cup V^i_{HJ}$;
    \EndFor \textbf{end for}
    \State \textbf{Adversarial RL Training}
    \State Initialize $\theta$; Environment $\mathcal{E}$  
    \While{not converged}
    \State $\tau_{\text{traj}} = \{ \}$
        \For {\text{rollout} $j=1 , \ldots, \textit{M}$}
        \State sample $i \sim B(1, N)$, construct $d^{i}$ based on $V^i_{HJ}$;
        \State $\tau_{j} \gets$ run policy $\pi_{\theta}$ and $d^i$ until termination;
        \State collect trajectory: $\tau_{\text{traj}} \gets \tau_{\text{traj}} \cup \tau_{j}$;
        \EndFor \textbf{end for}
        \State Update $\pi_{\theta} \gets \text{PPO \cite{schulman2017proximal}}( \tau_{\text{traj}} ,\pi_{\theta} )$
    \EndWhile \textbf{end while}
\end{algorithmic}
\end{algorithm}

%% file: sections/Numerical_Simulation.tex
\section{Numerical Simulations}

We evaluated \abbv\ in two simulated tasks: a one-vs-one reach-avoid game and quadrotor stabilization. 
The former involves a joint 4D dynamical system, which the OptimizedDP \cite{bui2022optimizeddp} can solve in the full state space with sufficient precision for the robust policy. 
The latter task involves a 12D quadrotor system that is computationally intractable for numerical HJ reachability solvers, where we focus on task-specific states to maintain computational accuracy. 
We demonstrate the consistency of the robust policy obtained from \abbv\ by comparing it with the HJ value function in the reach-avoid game, and it delivers a comparable performance to other robust adversarial RL baseline methods.

\subsection{One vs. One Reach-Avoid Game}
\label{RAG}
In the one vs. one reach-avoid game, the defender aims to capture the attacker while the attacker seeks to arrive at the destination without being captured. 
We follow a similar game pattern using the single integrator (SIG) dynamics to \cite{hu2023multi} but without obstacles. 
The SIG  dynamics are as follows:
\begin{equation}
\begin{array}{ll}
\dot{x}_{A}(t)=v_A u(t), \quad x_A(0) = x_{A0} \\
\dot{x}_{D}(t)=v_D d(t), \quad x_D(0) = x_{D0}
\end{array}
\label{SIG}
\end{equation}
where $x_{A}$ and $x_{D}$ are the 2D states that represent the positions of the attacker and the defender respectively, 
$x_{A0}$ and $x_{D0}$ are their initial positions, $v_A$ and $v_D$ are the constant speed of the attacker and the defender, 
in this game, we set $v_A=1.0$ and $v_D=1.5$ respectively, 
$u(t)$ and $d(t)$ are the control inputs of the attacker and the defender respectively at the time $t$. 
The target set $R^{11}$ and the avoid set $A^{11}$ for this one vs. one reach-avoid game are defined as follows:
\begin{equation}
\label{RA_11}
\begin{aligned}
R^{11}= & \left\{ x_{A} \in \mathcal{T} \right\} \cap \left\{\left\|x_{A}-x_D\right\|_2>r\right\} \\
A^{11} = & \left\{ \left\|x_{A}- x_D\right\|_2 \leq r\right\}
\end{aligned}
\end{equation}
where $\mathcal{T}$ is the destination to the attacker, 
and $r$ is the capture radius of the defender. 
Given these two sets and the dynamics defined in Eq.\eqref{SIG}, we can solve Eq. \ref{HJI} and obtain a 4D BRT $\mathcal{RA}^{11}_{\infty}(R^{11}, A^{11})$. 
The attacker will win if the initial joint state $(x_{A0}, x_{D0})$ lies within the $\mathcal{RA}^{11}_{\infty}(R^{11}, A^{11})$ using the optimal control \cite{chen2014multiplayer}:
\begin{equation}
u^*\left(x_A, x_D, t\right)=-v_A \frac{p_a\left(x_A, x_D,-t\right)}{\left\|p_a\left(x_A, x_D,-t\right)\right\|_2} 
\end{equation}
where $p_{a} = \frac{\partial V_{HJ}}{\partial x_{A}}$ is the partial derivative of the one vs. one HJ value function to the attacker. 
When the initial joint state $(x_{A0}, x_{D0})$ lies out of the $\mathcal{RA}^{11}_{\infty}(R^{11}, A^{11})$, the defender will guarantee to capture the attacker following the optimal control\cite{chen2014multiplayer}:
\begin{equation}
\label{1v1d}
d^*\left(x_A, x_D, t\right)=v_D \frac{p_d\left(x_A, x_D,-t\right)}{\left\|p_d\left(x_A, x_D,-t\right)\right\|_2} 
\end{equation}
where $p_{d} = \frac{\partial V_{HJ}}{\partial x_{D}}$ is the partial derivative of the one vs. one HJ value function to the defender. 

\abbv\ considers the optimal attacker policy from Eq.\eqref{1v1d} as an adversarial disturbance and aims to obtain a robust policy for the defender. 
In the adversary generation phase, we set the attacker and defender with the same control input range, $N=1$. 
To avoid the problem that the gradient of the one vs. one HJ value function is almost zero when the attacker lies outside of $\mathcal{RA}^{11}_{\infty}(R^{11}, A^{11})$ causing the attacker not to move, we also compute a one vs. zero HJ value function where only one attacker is in the game so that the attacker will continue moving to the destination when it lies outside of the $\mathcal{RA}^{11}_{\infty}(R^{11}, A^{11})$. 
Then, in the adversarial RL training stage, we collect a total of $10^7$ steps to train the defender policy. 
The reward function consists of three components: a bonus of 200 is awarded if the defender captures the attacker; a penalty of 200 is applied if the attacker reaches the destination; and the relative distance between the two is subtracted as an additional penalty. 
As for RARL\cite{pinto2017robust}, and RAP\cite{vinitsky2020robust}, the control policy of the attacker is represented by adversarial neural networks.
Finally, we compare the learned robust controllers trained through \abbv\ with RARL\cite{pinto2017robust} and RAP\cite{vinitsky2020robust}.

\begin{figure}[ht]
\centering
\includegraphics[width=0.48\textwidth]{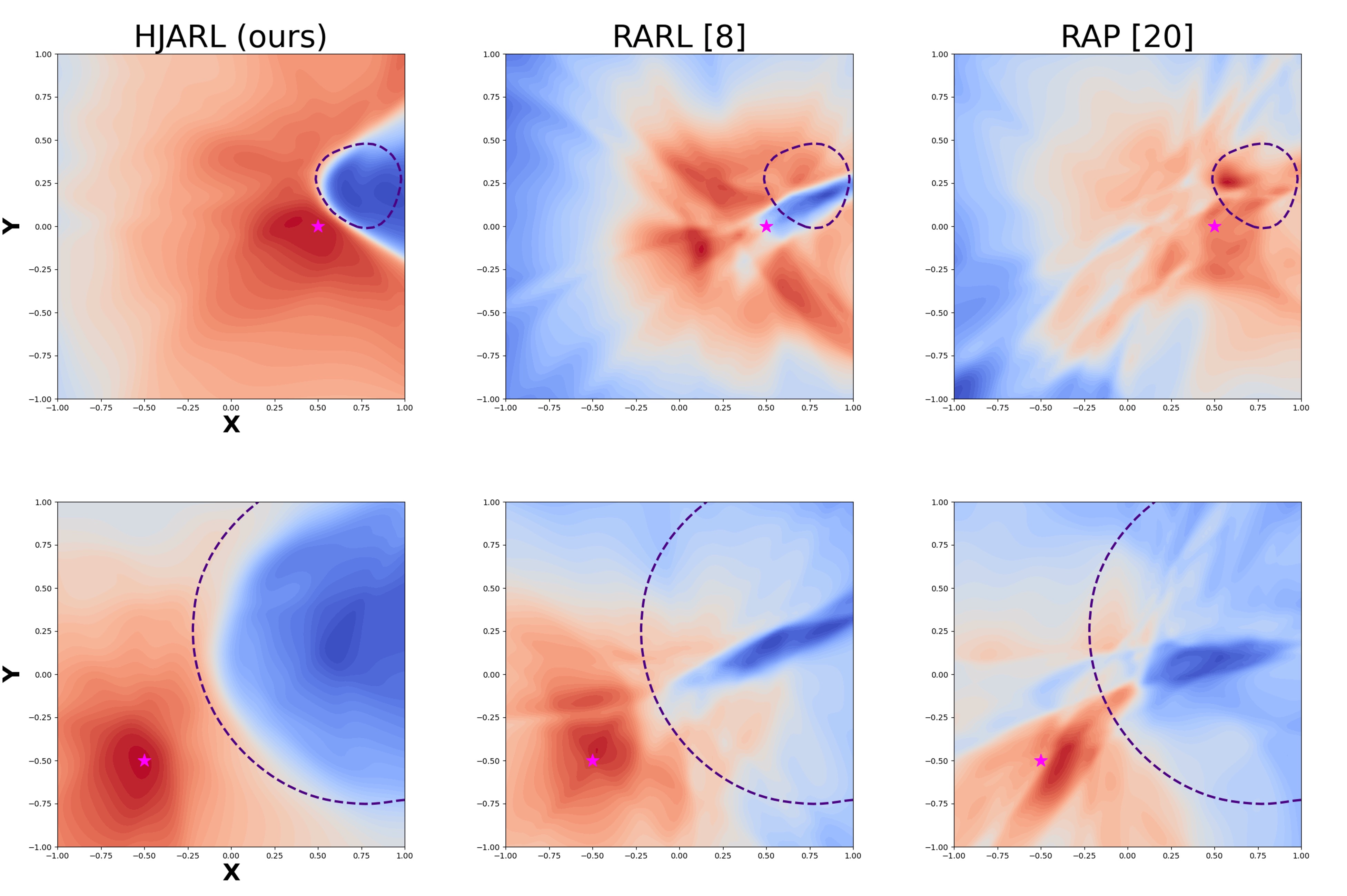}
\caption{Trained critic networks heatmaps and the zero-level $\mathcal{RA}^{11}_{\infty}(R^{11}, A^{11})$ (purple dash lines) with SIG dynamics. 
The first and the second rows show the values of the defender's initial positions at $[0.5,0.0]$ and $[-0.5, -0.5]$ respectively (magenta stars). 
}
\label{heatmaps}
\end{figure}

\textbf{Consistency to the HJ Value Function.} 
As illustrated in Fig.\ref{heatmaps}, \abbv\ demonstrates the strong consistency with the HJ value function through highly overlapping regions. 
Given that PPO is an actor-critic algorithm, we generate the heatmaps using the trained critic network in \abbv\ alongside the zero-level $\mathcal{RA}^{11}_{\infty}(R^{11}, A^{11})$ obtained from the HJ value function with two initial defender positions. 
In particular, regions with low values on the heat maps correspond closely to the regions enclosed by the zero-level $\mathcal{RA}^{11}_{\infty}(R^{11}, A^{11})$. 
When the attacker lies outside the zero-level $\mathcal{RA}^{11}_{\infty}(R^{11}, A^{11})$, the defender is guaranteed to capture the attacker under the optimal HJ control, regardless of the attacker's control policy. 
Hence, with this consistency, the trained critic network of \abbv\ can be used as an approximation to the HJ value function and work as a rough guaranteed attacker-winning region.
In addition, this trained critic network can help to check the degree of convergence of the training policy.
In contrast, although RARL and RAP also leverage Nash Equilibrium similar to the HJ reachability analysis, the dynamic nature of their adversaries results in an evolving MDP. 
This variability leads to discrepancies in their value functions, deviating from the consistent behavior exhibited by the true $\mathcal{RA}^{11}_{\infty}(R^{11}, A^{11})$. 

\begin{figure}[ht]
\centering
\includegraphics[width=0.48\textwidth]{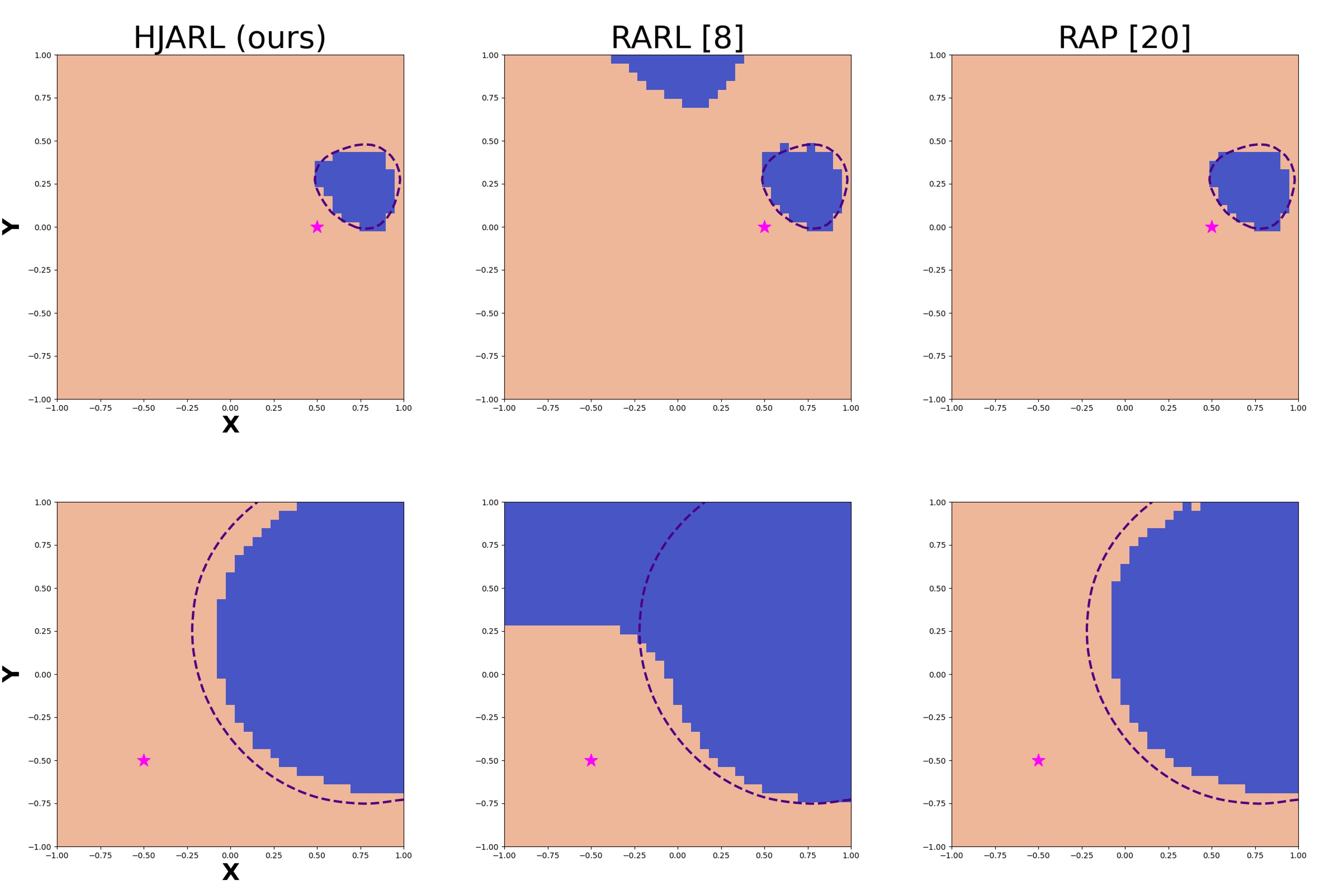}
\caption{Trained policy game performances and the zero-level $\mathcal{RA}^{11}_{\infty}(R^{11}, A^{11})$ (purple dash lines) with SIG dynamics. Initial attacker positions are uniformly generated across the map at intervals of 0.05 grid units, with the defender’s initial position fixed. 
The first and the second rows show the game results of the defender's initial positions at $[0.5,0.0]$ and $[-0.5, -0.5]$ respectively (magenta stars). 
}
\label{scores}
\end{figure}

\textbf{Comparable Performances.} 
As depicted in Fig.\ref{scores}, \abbv\ achieves comparable performances to the RAP method and outperforms the RARL method. 
We evaluated the performance of the trained policy networks in a batch of games. 
With the fixed initial defender position, we traverse the initial attacker positions across the entire map uniformly. 
If the defender successfully captures the attacker, the corresponding initial attacker's position is marked in peach; 
and if the attacker reaches the destination, its initial position is marked in blue. 
The capture performances indicate that both \abbv\ and RAP yield capture outcomes similar to those predicted by HJ reachability. 
In contrast, RARL fails to capture the attacker in certain areas where the defender should prevail. 
In addition, the capture performances match the heatmaps generated by the critic networks in \abbv, further strengthening the consistency of \abbv\ with the optimal HJ reachability policy.

\subsection{Quadrotor Stabilization}
The quadrotor is a representative example of a high-dimensional nonlinear system. 
Sabatino models a quadrotor as a 12D dynamical system \cite{sabatino2015quadrotor}.
However, due to the curse of dimensionality, we only focus on the main task-specific six of these dimensions, which we use to define a 6D nominal dynamical system for HJ reachability analysis. 
The nominal system is defined as follows:
\begin{equation}
\begin{array}{l}
\dot{\phi}=p+r \left( \cos{\phi} \frac{\sin{\theta}}{\cos{\theta}}\right)+ q \left( \sin{\phi} \frac{\sin{\theta}}{\cos{\theta}}\right) \\
\dot{\theta}= q \left( \cos{\phi} \right)-r \left( \sin{\phi} \right) \\
\dot{\psi}=r \frac{ \cos{\phi}}{\cos{\theta}} + q \frac{\sin{\phi}}{\cos{\theta}}\\
\dot{p}=\frac{I_y-I_z}{I_x} r q + \frac{u_x+d_{x}}{I_x} \\
\dot{q}=\frac{I_z-I_x}{I_y} p r + \frac{u_y+d_{y}}{I_y} \\
\dot{r}=\frac{I_x-I_y}{I_z} p q + \frac{u_z+d_{z}}{I_z}
\end{array}.
\end{equation}
where $\phi, \theta, \psi$ are Euler angles (roll, pitch, and yaw angles respectively) in the earth frame as shown in \cite{sabatino2015quadrotor}, 
$p, q, r$ are angular velocities (roll rate, pitch rate, and yaw rate respectively) in the body frame,
$u = [u_x, u_y, u_z]$ are control input torques generated by the differences among motors' speeds, 
$d = [d_{x}, d_{y}, d_{z}]$ are disturbance torques generated by disturbances like the wind.

External factors such as wind and the discrepancy between the true dynamics and the nominal system are modeled as disturbances. 
We assume that these disturbances can be applied to the quadrotor as actions. 
At the initial stage of our algorithm, we increase the upper bound of the disturbances to twice that of the control input’s upper bound and compute the value function $V^i_{HJ}$ at uniform intervals of 0.1, resulting in $N=21$ different value functions.  
In the second stage, we collect a total of $10^7$ steps to train the robust policy. 
The reward function comprises four components: a penalty proportional to the clipped action, scaled by $10^{-4}$; a penalty based on angular velocities, scaled by $10^{-3}$; a penalty of 100 for a quadrotor crash; and a penalty proportional to the distance between the current position and the destination. 
All baseline methods are trained with the same $10^7$ steps. 

\begin{table}[!htb]
\centering
\caption{Performances of Quadrotor on Episode Length}
\begin{tabular}{cccc}
\hline & Random HJ & Random & Constant \\
\hline \abbv\ (ours) & $\mathbf{702\pm 454}$ & $901 \pm 295$ & $\mathbf{677 \pm 457}$ \\
\hline PPO \cite{schulman2017proximal} & $287 \pm 430$ & $\mathbf{968\pm 171}$ & $638 \pm 474$ \\
\hline RARL \cite{pinto2017robust} & $253 \pm 408$ & $738 \pm 433$ & $313 \pm 449$ \\
\hline RAP \cite{vinitsky2020robust} & $257 \pm 411$ & $802 \pm 394$ & $675 \pm 458$ \\ 
\hline
\end{tabular}
\label{quadortor_performance}
\end{table}

\textbf{Performance Analysis.} 
The results in TABLE \ref{quadortor_performance} show that \abbv\ delivers strong performance across all evaluation conditions, comparable to other robust adversarial RL methods. 
We evaluated the performance of the algorithms in three environments characterized by different types of disturbances applied directly to the first two dimensions of the control actions: random HJ disturbance, random disturbance, and constant disturbance. 
Random HJ disturbance is generated by selecting a $V_{HJ}^i$ randomly from the value function buffer $\mathcal{V}$; 
random disturbances are sampled randomly at every step;
constant disturbances remain fixed throughout the entire episode. 
For each seed, we conducted 10 episodes (with a maximum of 1000 steps per episode), totaling 30 episodes per environment. 
To ensure a robust evaluation, we average the number of steps between three different seeds. 
The high standard deviation is attributed to the fragility of the quadrotor stabilization process, as it is prone to crashing if the control inputs are not properly designed. 
\abbv\ achieves comparable performance in these settings compared to other learning baselines; RAP and pure PPO perform well in random and constant disturbance environments; while RARL performs poorly. The inferior performance of RARL may be due to the adversary overfitting to the training distribution or getting stuck in local minima.

%% file: sections/Experiments.tex
\section{Real-World Experiments}
\begin{figure}[ht]
\centering
\includegraphics[width=0.45\textwidth]{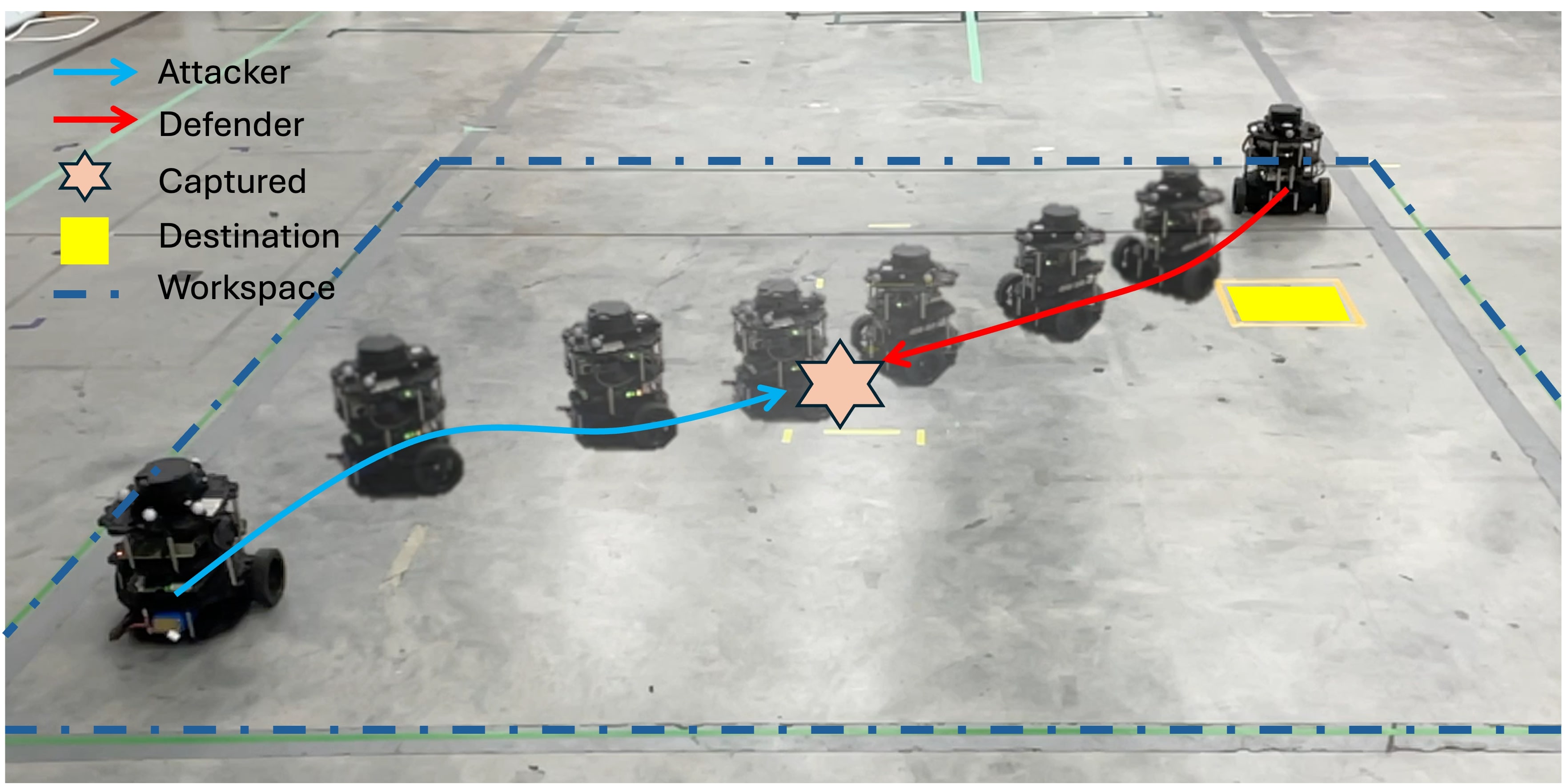}
\caption{The real-world one vs. one reach-avoid game with two TurtleBot3 Burger robots.}
\label{fig:game_real}
\end{figure}

We test \abbv\ in a real-world one vs. one reach-avoid game with two TurtleBot3 Burger robots. 
The dynamics of the TurtleBot3 Burger is a 3D DubinCar model \cite{he2023efficient}.
The game is carried out in a square arena $2m \times 2m$ with a square destination shown in Fig. \ref{fig:game_real}. 
We implement two different control policies for the attacker: HJ control and manual control. 
As for the HJ control, the full 6D HJ value function with all-time slices requires approximately 120 gigabytes and thus is impractical to implement.
Instead, we use a 3D HJ value function from a one vs. zero reach game for the attacker to generate control inputs to drive the attacker toward the destination. 
Regarding the defender policy, we train the defender’s control over $10^7$ steps using \abbv\ where the attacker uses the HJ control. 

\begin{figure}[ht]
\centering
\includegraphics[width=0.45\textwidth]{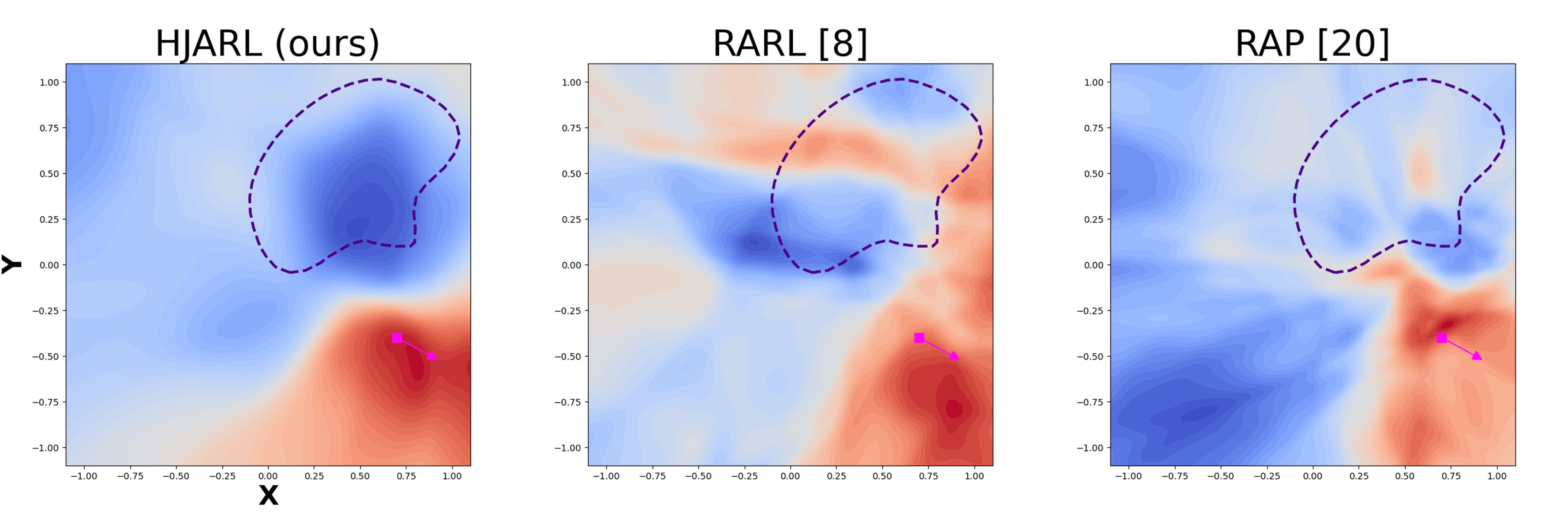}
\caption{Trained critic networks heatmaps and the zero-level HJ BRT (purple dash lines) with DubinCar model.
The initial defender is at $[0.7, -0.4, -0.5]$ with the arrow pointing in its direction (magenta square and arrow).}
\label{fig:dubin_real}
\end{figure}

As illustrated in Fig.\ref{fig:dubin_real}, \abbv\ still demonstrates the consistency with the HJ BRT, while baseline methods do not. Though the learned critic network does not perfectly overlap with the ground truth value function, it still provides more insights than the black-box learned value networks. 
As listed in TABLE \ref{real_world}, \abbv\ achieves the best performance in both attacker control policies. 
We conducted seven games with the defender and the attacker positioned at different locations on the map. 
In these scenarios, the relative positions fell outside the HJ BRT, indicating that the defender could capture the attacker with the optimal control policy. 
\abbv\ demonstrates superior performance with a HJ-controlled attacker, achieving a capture rate of $85.7\%$ (6/7), outperforming RARL and RAP, which achieved $57.1\%$ and $71.4\%$, respectively. 
Similarly, \abbv\ maintains a leading position with a manually-controlled attacker, reaching a $57.1\%$ capture rate, compared to $28.6\%$ for RARL and $ 42.9\%$ for RAP. 
These results underscore the robustness and efficiency of \abbv\ in both control settings.

\begin{table}[ht]
\centering
\caption{Performances of real-world experiments}
\begin{tabular}{ccc}
\hline
     & HJ attacker & Manual attacker \\ \hline
\abbv\ (ours) &  $\textbf{6}/7$ ($\textbf{85.7}\%$) &  $\textbf{4}/7$  ($\textbf{57.1}\%$)  \\ \hline
RARL \cite{pinto2017robust} &  $4/7$ ($57.1\%$)    & $2/7$  ($28.6\%$)  \\ \hline
RAP \cite{vinitsky2020robust}  & $5/7$ ($71.4\%$)   & $3/7$  ($42.9\%$)  \\ \hline
\end{tabular}
\label{real_world}
\end{table}

%% file: sections/Conclusion.tex
\section{Conclusion}
In this work, we introduce \abbv, a robust RL training framework with interpretable disturbance generation via HJ reachability analysis. 
Our approach leverages HJ value functions to create an interpretable disturbance generation for a robust policy training pipeline, where we evaluated across two simulation environments and one real-world experiment. 
We show that \abbv\ achieves robust performances comparable to state-of-the-art methods while retaining an interpretable adversary in terms of disturbance generation and physical explanation.
Despite the inherent challenges of scaling model-based methods to high-dimensional systems and ensuring robust guarantees, recent advances in learning-based techniques that approximate high-dimensional HJ reachability value functions could be employed in future work.

\newpage